\documentclass[11pt]{article}
\usepackage[margin=1in]{geometry}
\usepackage[T1]{fontenc}
\usepackage[utf8]{inputenc}
\usepackage{amsmath,amssymb}
\usepackage{graphicx}
\usepackage{booktabs}
\usepackage{makecell}
\usepackage{caption}
\usepackage{subcaption}
\usepackage{hyperref}
\usepackage{enumitem}
\usepackage{microtype}
\usepackage{parskip}
\title{Training a Large Language Model for Medical Coding Using Privacy-Preserving Synthetic Clinical Data}

\author{
\begin{tabular}{c}
John Cook\textsuperscript{1,$\ddagger$},
Michael Wyatt\textsuperscript{2,$\ddagger$},
Peng Wei\textsuperscript{1,$\ddagger$},
Iris Chin\textsuperscript{1},
Santosh Gupta\textsuperscript{1} \\[0.4em]
Van Zyl Van Vuuren\textsuperscript{1},
Richie Siburian\textsuperscript{1},
Amanda Spicer\textsuperscript{1},
Kristen Viviano\textsuperscript{1} \\[0.4em]
Alda Cami\textsuperscript{1},
Raunaq Malhotra\textsuperscript{1},
Zhewei Yao\textsuperscript{2},
Jeff Rasley\textsuperscript{2,$\dagger$},
Gaurav Kaushik\textsuperscript{1,*,$\dagger$} \\[0.7em]
\textsuperscript{1}Veradigm \quad
\textsuperscript{2}Snowflake \\[0.7em]
{\small *Corresponding author, \textsuperscript{$\dagger$}Co-senior authors, \textsuperscript{$\ddagger$}Equal contribution }
\end{tabular}
}

\date{}

\begin{document}
\maketitle

\begin{abstract}
Improving the accuracy and reliability of medical coding reduces clinician burnout and supports revenue cycle processes, freeing providers to focus more on patient care. However, automating the assignment of ICD-10-CM and CPT codes from clinical documentation remains a challenge due to heterogeneous records, nuanced coding guidelines, and long-tail distributions. Large language models have been proposed to help or automate specific medical coding tasks. However, foundation models are not explicitly trained for medical coding and zero-shot coding has yielded poor results. We investigate whether a modern open-weight foundation model can be adapted for an expert-level medical coding task using privacy-preserving synthetic training data derived from electronic health records. We fine-tune Llama 3-70B on pairs of clinical notes and gold codes generated from EHR-grounded templates and coding policies, then evaluate exact-code prediction for ICD-10-CM and CPT. A zero-shot baseline with the unadapted model achieved an F1 score of 0.18 for exact code match. After fine-tuning on the synthetic corpus, exact-match F1 exceeded 0.70, representing a large absolute gain across both code systems. Notably, performance remained high on complex categories that often require multi-step clinical reasoning and code composition, including Advanced Illness and Frailty classes, and the model retained its performance on medical comprehension tasks. These results indicate that synthetic, policy-aware data can efficiently teach a general-purpose large language model to support precise medical coding without exposing protected health information. The approach offers a practical path for training coding agents safely and iteratively on specific tasks that represent real-world populations.
\end{abstract}

\newpage

\tableofcontents
\newpage

\section{Introduction}

In the United States, healthcare providers must submit claims to insurers in order to receive reimbursement for services rendered. These claims require standardized clinical codes that classify patient conditions and services delivered. Accurate and reliable medical coding underpins not only the financial workflows of healthcare providers but also the generation of secondary data for clinical research, quality measurement, and care optimization. Clinical coding is a labor-intensive and error-prone process with direct implications for billing accuracy, revenue management, and the quality of downstream analytics. Prior work shows substantial variability in coding accuracy and significant operational burden on providers, reinforcing the importance of high-quality documentation and coding for both financial performance and secondary data use~\cite{venkatesh2023}.

Despite its centrality to healthcare operations, medical coding is time-consuming, inconsistent, and error-prone, as it relies on expert coders to interpret complex and heterogeneous clinical documentation. Observational and simulation studies report wide variation in clinical coding accuracy, with manual coding error rates spanning from about 50\% to nearly 100\% accuracy and median performance near 80\% in some settings. This variability reflects undercoding, miscoding, and challenges translating complex clinical narratives into standardized diagnosis and procedure codes, underscoring the difficulty coders face in consistently applying ICD and related code sets~\cite{venkatesh2023,omalley2005}. The challenge is compounded by variability in clinical note styles, evolving coding standards, and the need to process large volumes of claims expeditiously. Additionally, medical coding is a complex and multi-stage process spanning documentation review, code abstraction, validation against payer and regulatory rules, quality auditing, and downstream claim edits. Each stage introduces latency and the potential for error, particularly when steps are performed by different roles using fragmented tools and inconsistent reference standards. Resulting errors in coding delay reimbursement, increase denial rates, create compliance risks, and propagate into financial and clinical analytics.

Machine-assisted medical coding offers the potential to improve efficiency, consistency, and scalability in clinical documentation and billing workflows. Early systems demonstrated feasibility but were narrowly scoped and struggled to generalize across complex coding tasks and settings~\cite{stanfill2010}. Subsequent reviews published between 2010 and 2022 document a shift from rule-based and classical machine learning approaches toward deep neural architectures for ICD diagnosis and procedure coding, while consistently noting unresolved challenges, including sparse performance on rare codes, limited external validation, and a lack of evidence for sustained real-world deployment beyond curated research datasets~\cite{almeida2022,amato2022,ji2024,kaur2025}.

Recent advances in large language models (LLMs) represent a potential inflection point for clinical text understanding. LLMs can be continuously improved through domain-specific fine-tuning and further optimized using human feedback and alignment techniques~\cite{ouyang2022,bai2022}. When scaled and adapted to medical domains, LLMs can capture nuanced context from unstructured clinical narratives and have demonstrated near-clinician performance on a range of medical question-answering and reasoning benchmarks, indicating that a single foundation model can generalize across multiple documentation-intensive tasks~\cite{singhal2023,singhal2025}.

However, these results do not directly establish readiness for automated medical coding, and the optimal application of LLMs to structured billing tasks remains an active area of investigation. Evaluations of base and lightly adapted LLMs report low exact-match performance and frequent generation of invalid or inappropriate codes when models are applied directly to coding tasks without task-specific constraints~\cite{soroush2024}. Recent benchmarks across ICD-9, ICD-10, and CPT settings commonly observe exact-match accuracies in the 30-45\% range, with systematic hallucination of non-existent or semantically mismatched codes in the absence of external tools, retrieval, or rule-based guardrails~\cite{lee2024}.

Retrieval-augmented and tool-assisted coding systems demonstrate substantial performance gains relative to base LLM prompting. Recent work evaluating retrieval-augmented large language models for medical coding confirms potential accuracy gains over base prompting, but also underscores that such improvements are task-dependent and do not uniformly translate across coding scenarios~\cite{klang2025}. In narrowly scoped settings, large language models augmented with explicit lookup tools and code descriptions have reported near-perfect accuracy on constrained single-term ICD-10-CM tasks~\cite{kwan2024tools}. More generally, lookup-before-coding and retrieval-grounded pipelines that condition generation on code definitions and historical examples have been shown to match or exceed human coder performance on focused emergency department datasets and to achieve very high accuracy on single-condition tasks under retrospective evaluation~\cite{kwan2024lookup}. These gains, however, can come at the cost of increased system complexity and reliance on carefully curated retrieval corpora, which can limit immediate generalization to broad, real-world coding workloads.

Domain-specific fine-tuning of large language models can also improve performance on controlled variation tasks and discharge summaries, although exact-match accuracy remains imperfect~\cite{hou2025}. Fine-tuned models incorporating specialized ICD-10 knowledge show substantial gains over zero-shot prompting and traditional deep-learning baselines, yet continue to exhibit non-trivial error rates, particularly for complex encounters and low-prevalence codes. Hybrid frameworks that combine entity extraction, retrieval, and re-ranking report strong disease extraction performance (F1 $\approx$ 0.83) and improved ICD-10 prediction accuracy (micro-F1 $\approx$ 0.60) in benchmark settings, but precision and consistency degrade under more realistic annotation conditions~\cite{puts2025,dasbaksi2024}. Early pilot deployments of retrieval-augmented ICD-10 assistants in production coding workflows demonstrate improvements in lead-term identification and coder efficiency, while underscoring residual variability, sensitivity to documentation style, and the continued need for human review prior to claim submission.

Together, these studies suggest that retrieval, hybrid architectures, and fine-tuning for controlled domains are potential strategies to improve on base LLMs. However, important gaps remain in our understanding of how well these methods scale to full clinical notes, multi-code encounters, rare or low-frequency codes, and diverse documentation styles. Moreover, it remains unknown whether fine-tuned models alone can match the accuracy of more complex hybrid systems.

In this work, we address these gaps by investigating whether a modern open-weight foundation model can be adapted for expert-level medical coding. We fine-tune Llama-3.3-70B-Instruct for ICD-10-CM and CPT assignment using privacy-preserving synthetic training data derived from EHR-grounded templates and clinical coding guidelines. Synthetic, policy-aware clinical text has emerged as a promising approach for mitigating data scarcity and privacy constraints in health NLP, enabling large-scale model training while reducing reliance on raw protected health information and complementing de-identification methods that may remain susceptible to re-identification risk~\cite{libbi2021,sarkar2024,smolyak2024,alshaikhdeeb2025}.

Our evaluation measures exact code prediction as well as categorical placement across diverse coding domains. A zero-shot baseline using the unadapted foundation model establishes a performance floor, against which we quantify gains from iterative fine-tuning. We report accuracy improvements across multiple code families, including challenging categories that require multi-step clinical reasoning. Finally, we assess whether domain adaptation preserves general medical comprehension, an essential requirement for real-world deployment. Through this analysis, we aim to characterize both the capabilities and limitations of fine-tuned large language models for medical coding under privacy-preserving constraints.

\section{Methods}

\subsubsection{Model and training setup}

We fine-tuned Llama-3-70B-Instruct using supervised fine-tuning on paired clinical notes and target medical codes. The coding task was formulated as structured text generation, with model outputs constrained to a predefined JSON schema to promote syntactic validity and facilitate downstream parsing.

For ICD-10-CM coding, each output record included the predicted code, a clinical rationale, and the associated evidence and localization that supports the assignment. For CPT coding, outputs included the predicted code and a corresponding clinical rationale. Training data were split into training and evaluation partitions using a fixed 95/5 split, with splits preserved across experiments to prevent leakage.

Model inputs were tokenized using the native Llama tokenizer with a maximum sequence length of 8,192 tokens. Full-parameter fine-tuning was performed using the ArcticTraining framework with ZeRO-3 optimization, optimizer state offloading, and FP16 precision. Training used the Adam optimizer with a learning rate of 1e-5, weight decay of 0.1, and a linear learning rate scheduler. Training proceeded for four epochs with a micro-batch size of four samples per device on eight H200 GPUs. An overview of the training workflow, including prompt formatting, data augmentation, and sequence packing, is shown in Figure 4.

\subsubsection{Synthetic data generation}

To enable large-scale training while minimizing privacy risk, all training and evaluation data were synthetically generated. Large language models are known to memorize portions of their training data, creating potential privacy risks when trained on real clinical notes. Because direct exposure to protected health information may be unwanted in production healthcare settings, we designed a controlled synthetic data generation pipeline that produces clinically realistic documentation without including any protected health information. Synthetic data generation was performed separately for ICD-10-CM and CPT coding due to differences in code structure and clinical usage.

\subsubsection{ICD-10-CM synthetic chart generation}

Synthetic ICD-10-CM charts were generated using a two-phase pipeline consisting of chart synthesis followed by evidence-linked labeling.

\textbf{Phase 1: synthetic chart generation}\\First, real clinical documents were collected in a secure environment and used to derive abstract meta-descriptions capturing clinical structure and provider documentation style while explicitly omitting patient-specific details. An ICD-10-CM seed code was randomly selected and used to anchor the synthesis process. Additional clinically plausible co-occurring ICD codes were then generated, and a language model produced a synthetic clinical note conditioned on the meta-description and code set. This process was repeated across diverse source documents to capture variability in content and style without retaining any original patient content.

\textbf{Phase 2: ICD-10-CM labeling}

Each synthetic chart was embedded and segmented into line-level units. ICD-10-CM codes and their descriptions were embedded and stored in an indexed code database. For each chart segment, semantic retrieval identified candidate codes, and a language model selected the most appropriate codes with explicit evidence attribution. Outputs were reviewed to account for evolving ICD terminology and newly suggested valid codes. This process was repeated across targeted clinical domains, including Advanced Illness (Adv), Frailty (Fra), and Social Determinants of Health (SDoH).

\subsubsection{CPT synthetic chart generation}

Synthetic CPT charts were generated using a protocol adapted for procedural coding. A CPT seed code was sampled from specialty-specific ranges within the official CPT catalog. A language model then generated a clinically consistent procedure note conditioned on the seed code and description.

To identify additional relevant procedures, the system generated short procedural descriptions rather than raw CPT codes, which are prone to hallucination in generative models. These descriptions were embedded and matched against an indexed CPT catalog using cosine similarity to retrieve valid, current codes. The top-N retrieved codes were evaluated first, with fallback expansion applied when no suitable match was identified. A constrained language model then selected the final CPT code set from valid candidates. Notes for which no valid CPT code could be resolved after this process were excluded during data generation.

All intermediate artifacts, including generated notes, candidate codes, similarity scores, and discarded samples, were retained for auditability and later analysis.

\subsubsection{Training data augmentation and packing}

To enhance the robustness of the ICD-10-CM coding predictions, we applied difficulty-based data augmentation to 30\% of the ICD-10-CM training set. Pairs of clinical notes were concatenated into longer composite samples, requiring the model to reason across multiple cases in a single context. Corresponding label sets were merged, and "line index" values were adjusted to account for the extended length of the concatenated clinical notes. This augmentation increased contextual difficulty and more closely reflected multi-problem clinical encounters.

Given that clinical notes can vary in length, we implemented sequence packing to increase training efficiency. During data loading and after the data augmentation described above, multiple independent note-code pairs were concatenated into a single packed sequence whenever their combined token length fit within the 8,192 token limit. Each sample was delimited between other samples in the final packed sequence, and position IDs were reset for each example to preserve token-level coherence.

This method reduces per-sample padding overhead, thereby increasing the proportion of useful tokens processed in each training step without altering the learning objective or token distribution. Importantly, data packing is distinct from the difficulty-based augmentation described above: packing serves as an efficiency optimization, whereas augmentation deliberately introduces multi-case reasoning challenges.

\subsubsection{Evaluation datasets}

Evaluation datasets consisted of held-out synthetic clinical charts with gold-standard ICD-10-CM or CPT annotations. Splits were fixed prior to training and preserved across experiments to prevent leakage. Near-duplicate documents were removed using text similarity thresholds.

\subsubsection{ICD-10-CM coding evaluation}

Model performance was evaluated at the clinical note level and aggregated across the evaluation set, with primary emphasis on diagnostic granularity and evidence attribution. \\ We assessed ICD-10-CM performance across increasing levels of diagnostic specificity, corresponding to category-level diagnosis (Level 0), subcategory-level diagnosis (Level 1), fine-grained diagnosis (Level 2), exact ICD-10 code assignment (Level 3), and exact ICD-10 code assignment with supporting evidence localization (Level 4). Predicted and reference codes were mapped to category, block, and chapter levels using standard ICD-10 taxonomies, and performance was computed independently at each level to characterize degradation with increasing granularity.\\\\Performance was measured using multiple complementary metrics:

\textbf{Exact matching}, requiring both the ICD-10-CM code and its evidence location to match the reference annotation.

\textbf{Code-only matching}, evaluating diagnostic accuracy without considering evidence localization.

\textbf{Evidence localization}, measured using Jaccard similarity between predicted and reference evidence spans.

\textbf{Response quality}, quantified as the proportion of charts yielding no valid predictions.

To identify systematic failure modes, outlier detection was performed using the interquartile range method, with analysis restricted to categories appearing in at least ten evaluation samples. To assess generalization beyond frequently observed labels, model performance was additionally analyzed as a function of training set frequency.

\subsubsection{CPT coding evaluation}

CPT coding performance was evaluated using exact set matching between predicted and gold codes. Precision, recall, and F1 score were computed at the note level. The CPT catalog version used during evaluation matched that used during data generation to prevent catalog drift. Dataset-level diagnostics produced during data generation, including code frequency, labels per document, and discard rates due to unresolved matches, were recorded to characterize dataset coverage. These diagnostics were used solely to assess dataset properties and were not used as inference-time performance metrics.

\subsubsection{Clinical expert review protocol}

A subset of synthetic charts was reviewed by clinical experts to establish an expert-validated reference set under controlled conditions. For each chart, experts examined the ICD-10-CM labels produced by our labeling process and approved or rejected each label and documented reasons for rejection. The accepted labels constituted the ground truth used to assess the model performance. This expert review was conducted to establish a clinically grounded benchmark for model performance, as automated metrics alone may not fully capture the accuracy and relevance of code assignments in real-world clinical contexts.

\section{Results}

\subsubsection{Overall ICD-10-CM and CPT performance}

We evaluated overall coding performance of the fine-tuned Llama 3-70B model on held-out synthetic datasets for ICD-10-CM and CPT assignment across multiple levels of diagnostic granularity. For ICD-10-CM, performance was evaluated from coarse category identification (Level 0) through subcategory-level diagnosis (Level 1), fine-grained diagnosis (Level 2), exact ICD-10 code assignment (Level 3), and exact ICD-10 code assignment with supporting evidence localization (Level 4); CPT performance was evaluated using exact CPT code matching.

The fine-tuned model substantially outperformed the zero-shot baseline across both ICD-10-CM and CPT code systems. Exact ICD-10 code matching (Level 3) achieved an F1 score of 0.704, representing an absolute improvement of 0.524 points over the baseline (Figure 1a; baseline F1 = 0.180). Comparable improvements were observed for CPT coding (Figure 1b), with the fine-tuned model achieving an overall F1 score of 0.736 compared with 0.193 for the baseline, representing a 0.543-point improvement.

When examining performance across ICD-10-CM code hierarchical levels (Figure 2), the highest accuracy was observed at the category-level (Level 0; F1 = 0.864). Performance declined gradually with increasing diagnostic specificity. The most stringent task, exact ICD-10 code matching with evidence localization (Level 4), achieved an overall F1 score of 0.629, with no abrupt drops observed across hierarchical levels.

Taken together, these results demonstrate that supervised fine-tuning on synthetic, policy-aware clinical data materially improves end-to-end medical coding accuracy across both diagnostic and procedural domains.

\subsubsection{Performance by complex clinical domain}

We next examined performance across three clinically relevant domains: Advanced Illness, Frailty, and Social Determinants of Health (SDoH). As shown in Figure 3A and summarized in Table 1, the fine-tuned model achieved strong performance across all three domains, substantially exceeding the overall zero-shot baseline performance (F1=0.18). \\ Performance was highest for Frailty (F1= 0.873), followed closely by Advanced Illness (F1 = 0.863). These domains are characterized by relatively explicit clinical documentation and well-defined diagnostic patterns, which likely contribute to more reliable code assignment.\\ Performance on SDoH-related codes was lower (F1 = 0.767), representing an absolute gap of approximately 10 percentage points relative to Advanced Illness and Frailty. Despite this gap, SDoH performance remained well above the overall baseline, indicating that the model captures meaningful social and contextual signals even in domains where documentation is often implicit, fragmented, or inconsistently recorded.

\subsubsection{Error patterns and low-performance code groups}

To characterize model limitations, we analyzed error patterns across ICD-10-CM categories and diagnostic groupings. As shown in Figure 3B, errors were concentrated in a small subset of category-level codes, particularly those related to psychosocial circumstances, physical environment, and functional limitations. At higher levels of aggregation, performance was comparatively stable. Chapter-level analysis (Figure 3C) revealed no extreme outliers, indicating that performance degradation is driven by semantic ambiguity at the category level rather than broad structural differences across ICD-10 chapters.

\subsubsection{Label frequency and model performance}

We analyzed the relationship between ICD-10-CM code prevalence in the training data and evaluation performance. As shown in Figure 5, low-frequency categories exhibited substantial variance in F1 score, indicating that prevalence alone is insufficient to guarantee accurate prediction. At the same time, categories with higher training frequency consistently achieved strong performance, with F1 scores clustering near the upper range. This pattern reflects a frequency threshold effect rather than a linear relationship, where sufficient representation stabilizes performance without ensuring continued gains. Remaining errors among low-frequency codes are therefore more likely attributable to documentation ambiguity than to data scarcity alone.

\subsubsection{Human expert evaluation}

To assess clinical validity beyond automated metrics, clinical experts reviewed 100 synthetic charts across the three clinical domains (Advanced Illness, Frailty, and Social Determinants of Health). For each chart, experts accepted or rejected the ICD-10-CM labels produced by the labeling process; the accepted labels constituted the expert-validated ground truth for evaluation. After applying a post-processing step to filter predictions to only SDoH, Frailty, and Advanced Illness-related codes, the fine-tuned model achieved an overall F1 of 0.44 (Figure 6 and Table 3). While the F1 scores reflect room for improvement (driven largely by lower precision), the model demonstrated strong recall across all ICD-10 hierarchy levels: 0.93 at Level 0 to 0.86 at Level 3. Precision remained comparatively lower, ranging from 0.4 (Level 0) to 0.32 (Levels 2 and 3), indicating that the model tends to generate additional codes beyond the expert-validated target set.\\\\Preservation of general medical knowledge after fine-tuning

We evaluated whether domain-specific fine-tuning for medical coding affected general medical knowledge using standard medical question-answering and reasoning benchmarks. As shown in Table 2, fine-tuning resulted in modest but consistent declines across several benchmarks, with no evidence of catastrophic degradation or collapse in performance. Accuracy remained high across all evaluated domains, indicating that specialization for structured coding tasks introduces bounded tradeoffs in general medical reasoning rather than wholesale loss of underlying knowledge.

\section{Discussion}

\subsubsection{Principle findings}

In this study, we show that a modern open-weight foundation model can be adapted for ICD-10-CM and CPT coding using privacy-preserving synthetic training data. Fine-tuning on synthetic, policy-aware clinical text substantially improves coding accuracy relative to a zero-shot baseline, with stable performance across increasing levels of diagnostic specificity. Performance remains strong for clinically explicit domains such as Advanced Illness and Frailty. However, we see lagging for codes related to Social Determinants of Health, which rely on implicit and inconsistently-documented information. This gap highlights the limits of single-pass prediction and underscores the importance of incorporating domain-specific reasoning, contextual validation, and human judgment in complex coding scenarios. Importantly, specialization for medical coding did not result in catastrophic degradation of general medical knowledge, as measured on standard medical comprehension benchmarks. These findings indicate that synthetic data can support meaningful progress in automated medical coding while mitigating privacy risks inherent in training on raw clinical notes.

These findings contrast with prior evaluations reporting poor medical coding performance for base large language models when applied without task-specific adaptation or constraints, and suggest that targeted fine-tuning and data design can meaningfully alter this conclusion~\cite{soroush2024}. Language models at that time, and given the conditions of the study, had performance was too poor for use in real-world settings. In contrast, our results suggest that fine-tuned models can be used appropriately in real-world medical coding workflows, while also clarifying the role such systems are likely to play in practice. At current performance levels, they support an AI-augmented coding paradigm in which foundation models function as coordinated assistants that reduce cognitive load, surface ambiguities, and support coders and compliance experts in oversight, exception handling, and final adjudication, rather than as fully autonomous replacements.

\subsubsection{Synthetic data as a privacy-risk mitigation strategy}

Large language models are known to memorize portions of their training data, creating potential privacy risks when trained directly on clinical documentation~\cite{carlini2023}. Because such risks are unacceptable in production healthcare environments, we designed a framework that uses fully synthetic clinical text to decouple model training from direct exposure to protected health information.

Our approach generates entirely synthetic notes grounded in real-world clinical structure and coding policy, rather than preserving original documents through de-identification. This design prioritizes population-level fidelity over record-level replication, which is appropriate for training and evaluation but may introduce distributional differences relative to real-world clinical data. Synthetic data should therefore be viewed as a risk-mitigation and early-validation strategy, not a substitute for validation on real clinical documentation under appropriate governance.

\subsubsection{Strengths and limitations of the synthetic data approach}

The primary strength of the synthetic data framework is control. It enables scalable training, explicit incorporation of coding guidelines, and targeted generation of clinically meaningful edge cases without exposing PHI. Evidence-linked labeling further aligns supervision with real-world coding requirements.

At the same time, important limitations remain. Synthetic notes may not capture the full variability, noise, and idiosyncratic documentation practices observed in operational EHRs. The current pipeline relies primarily on prompt-based controls to suppress PHI, which could be strengthened through additional automated validation steps such as deterministic filters or classification-based detection. In addition, constrained diversity of source records limits representativeness, underscoring the need for cautious interpretation of performance metrics derived from synthetic data. Continuous validation may also be necessary to ensure that performance is maintained as coding guidelines change.

\subsubsection{Interpreting hierarchical and domain-specific performance}

A notable finding of this work is the fine-tuned model's ability to capture diagnostic intent across increasing levels of granularity. Performance remains stable from category-level diagnosis through fine-grained classification, with degradation occurring primarily at the level of exact code assignment. This pattern suggests that the model learns hierarchical and semantic relationships among diagnostic concepts even when precise code binding becomes more challenging.

This hierarchical stability is clinically meaningful. In many coding workflows, high-level categorization and identification of relevant diagnostic families precede final code resolution and audit. The absence of sharp performance drops across levels of specificity indicates that large language models can support upstream clinical reasoning steps, while highlighting that exact code selection remains the principal bottleneck. The consistency of this pattern across domains further suggests that limitations at the finest level of granularity reflect representational challenges rather than insufficient model capacity.

Domain-specific performance differences reinforce this interpretation. Strong results for Advanced Illness and Frailty align with relatively explicit documentation patterns, whereas lower performance for SDoH reflects the implicit, context-dependent nature of these codes. Together, these findings emphasize documentation clarity and semantic grounding as primary drivers of exact-code accuracy, rather than code frequency alone.

\subsubsection{CPT coding considerations}

CPT coding presents distinct challenges relative to ICD-10-CM due to its procedural focus and lack of hierarchical structure. Unlike diagnostic codes, CPT codes often require precise alignment between free-text procedural descriptions and standardized billing terminology. Our results suggest that models readily learn procedural intent but may struggle to consistently bind that intent to exact CPT identifiers in the absence of explicit grounding.

These observations motivate a design approach that separates semantic understanding from final code selection. Grounding model outputs in structured code descriptions or authoritative catalogs may be particularly important for procedural coding, where small lexical differences can correspond to materially different billing outcomes. In professional coding workflows, recall may be at least as important as precision, as omitted procedure codes can have downstream financial and compliance implications. Accordingly, conservative design choices that prioritize auditability, such as constrained code selection and explicit grounding against curated catalogs, remain essential. Systematic evaluation of description-grounded approaches and their impact on recall-precision tradeoffs represents an important direction for future work.

\subsubsection{Human evaluation and implications for clinical workflows}

Automated performance metrics alone are insufficient to assess suitability for real-world deployment. Human expert review remains essential for validating clinical appropriateness, evidence support, and trustworthiness of code assignments. In this study, expert feedback highlighted that model predictions frequently captured the correct clinical intent, even when exact code selection was imperfect, underscoring the distinction between semantic understanding and billing-level correctness. The high recall observed across all ICD-10 hierarchy levels suggests that the model is effective at identifying clinically relevant concepts present in the chart, while the lower precision reflects a tendency to over-generate candidate codes. From a workflow perspective, this distinction is critical. Systems that surface relevant candidate codes with high recall may support human decision-making, even when final code selection requires expert judgment. Future refinement efforts should therefore prioritize improving precision to reduce the review burden on clinicians. However, this study was not designed to measure productivity, time-to-code, or user satisfaction, and no conclusions about operational efficiency should be drawn. Future evaluations should explicitly assess human-model collaboration under realistic workflow conditions.

\subsubsection{Statistics}

In this study, we report F1 scores aggregated at the document level, reflecting performance across the full set of diagnosis and procedure codes assigned to each clinical note. This approach emphasizes end-to-end coding accuracy at the encounter level and aligns with real-world billing workflows, in which codes are evaluated collectively rather than in isolation.

\subsubsection{Limitations \& Future Directions}

Limitations of this study include constrained source diversity, a limited scale of human evaluation, and training under a fixed, pre-determined compute budget, which may have capped achievable performance. Baseline comparisons were limited to a zero-shot model and did not include domain-specific baselines. Future work could evaluate these models in real-world clinical and billing settings under appropriate governance and human oversight, including multi-agent and human-in-the-loop workflows that reflect how coding decisions are made in practice. Importantly, such evaluations should prioritize downstream outcomes for provider practices, such as reduced rework, denial rates, and time-to-bill, rather than coding accuracy alone, which can be subjective and context-dependent.

\section{Figures}

\begin{figure}[ht]
\centering
\includegraphics[width=0.9\linewidth]{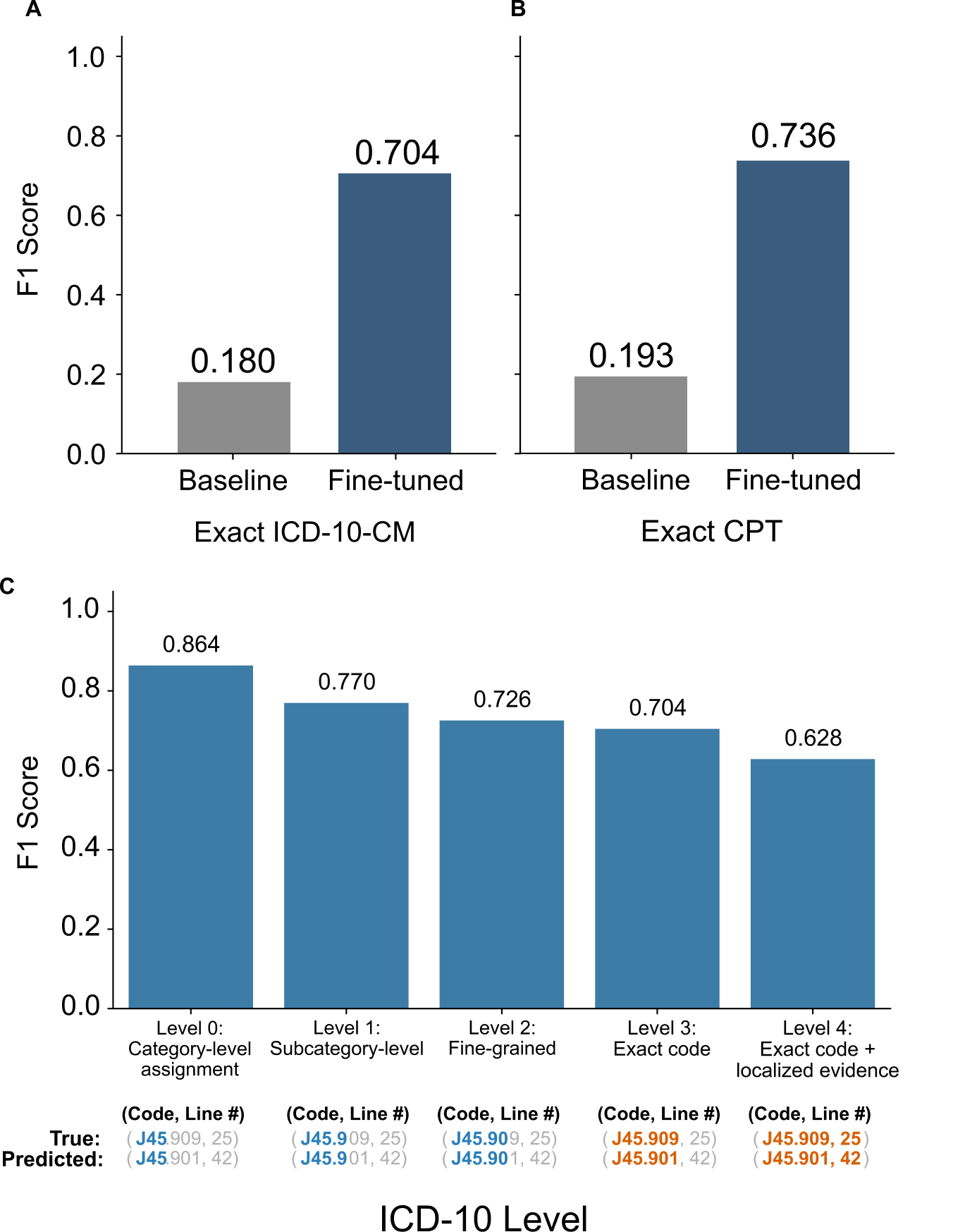}
\caption{Figure 1. Exact-match ICD-10-CM and CPT coding. F1 scores for the fine-tuned Llama-3-70B model evaluated on held-out synthetic (A) ICD-10-CM and (B) CPT datasets. Results reflect baseline inference without prompt engineering or semantic retrieval assistance.}
\end{figure}

\begin{figure}[ht]
\centering
\includegraphics[width=0.9\linewidth]{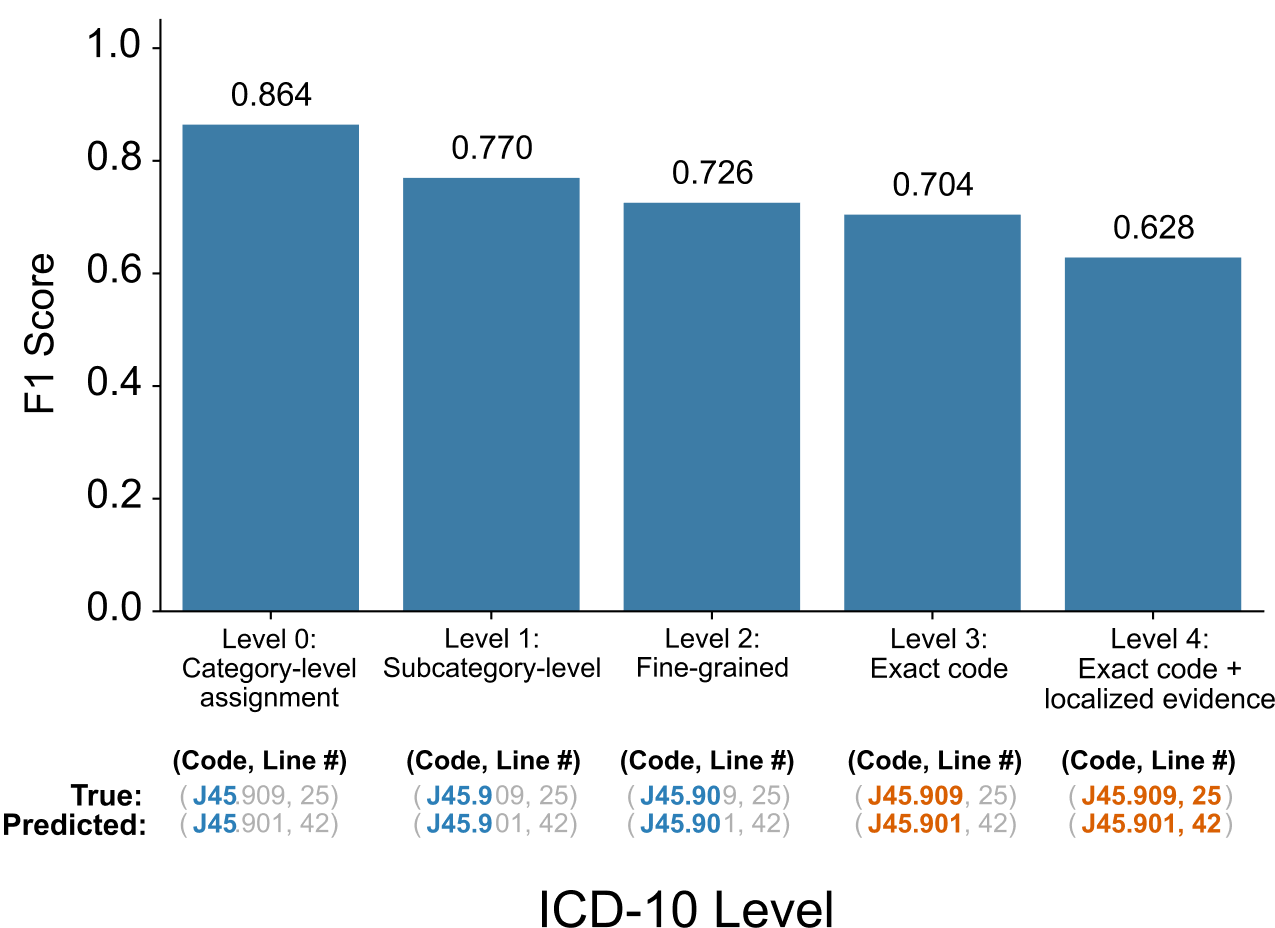}
\caption{Figure 2. ICD-10-CM coding performance across hierarchical levels. ICD-10-CM performance is reported at increasing levels of diagnostic specificity, from coarse category identification to exact ICD-10 code (Level 3) and exact code with supporting evidence attribution (Level 4). Performance is highest at the coarsest level and declines gradually as diagnostic specificity increases. CPT coding performance, evaluated independently using exact set matching, achieves an F1 score of 0.736. Results reflect baseline inference without prompt engineering or semantic retrieval assistance.}
\end{figure}

\begin{figure}[ht]
\centering
\begin{subfigure}{0.32\linewidth}
\centering
\includegraphics[width=\linewidth]{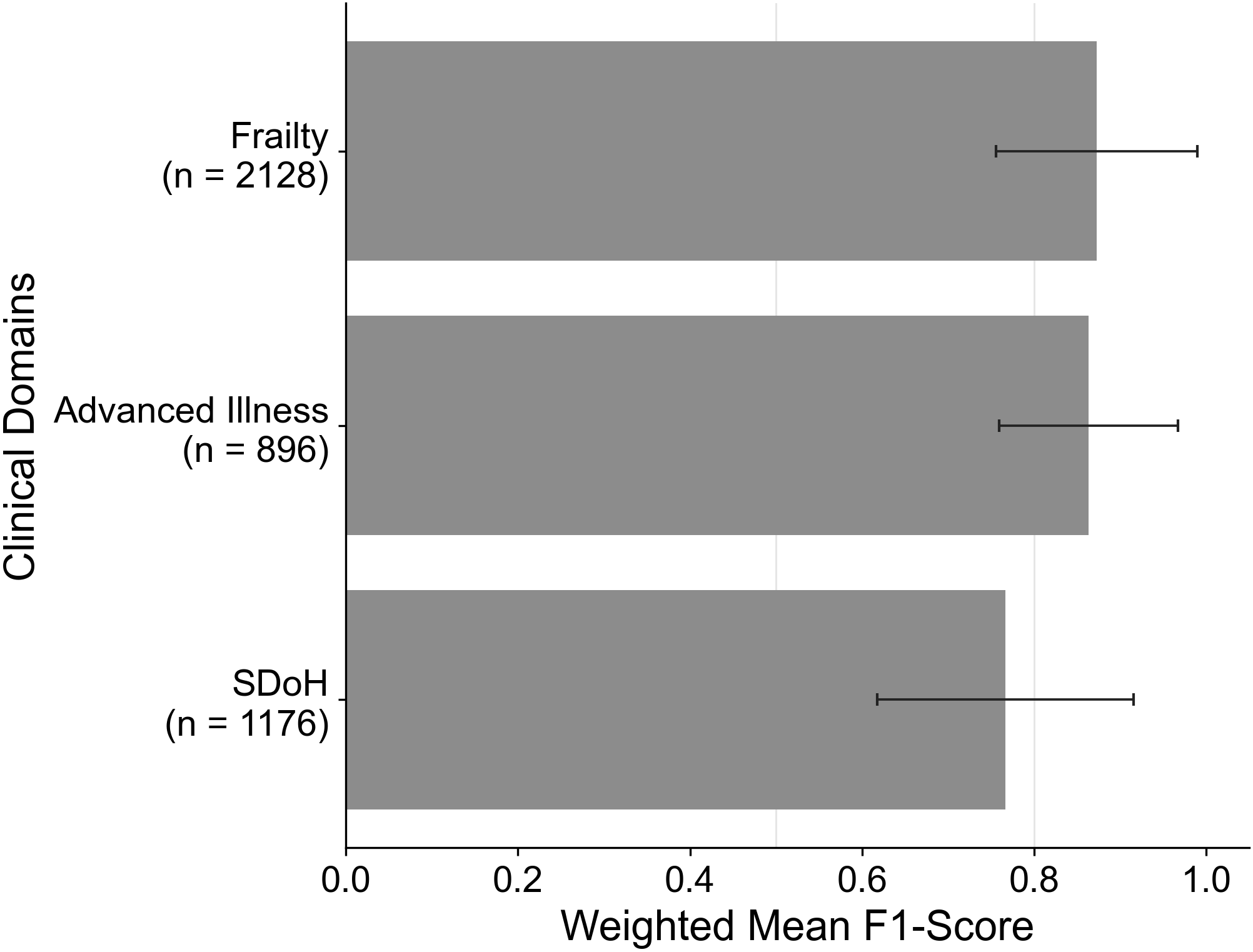}
\caption{}
\end{subfigure}
\begin{subfigure}{0.32\linewidth}
\centering
\includegraphics[width=\linewidth]{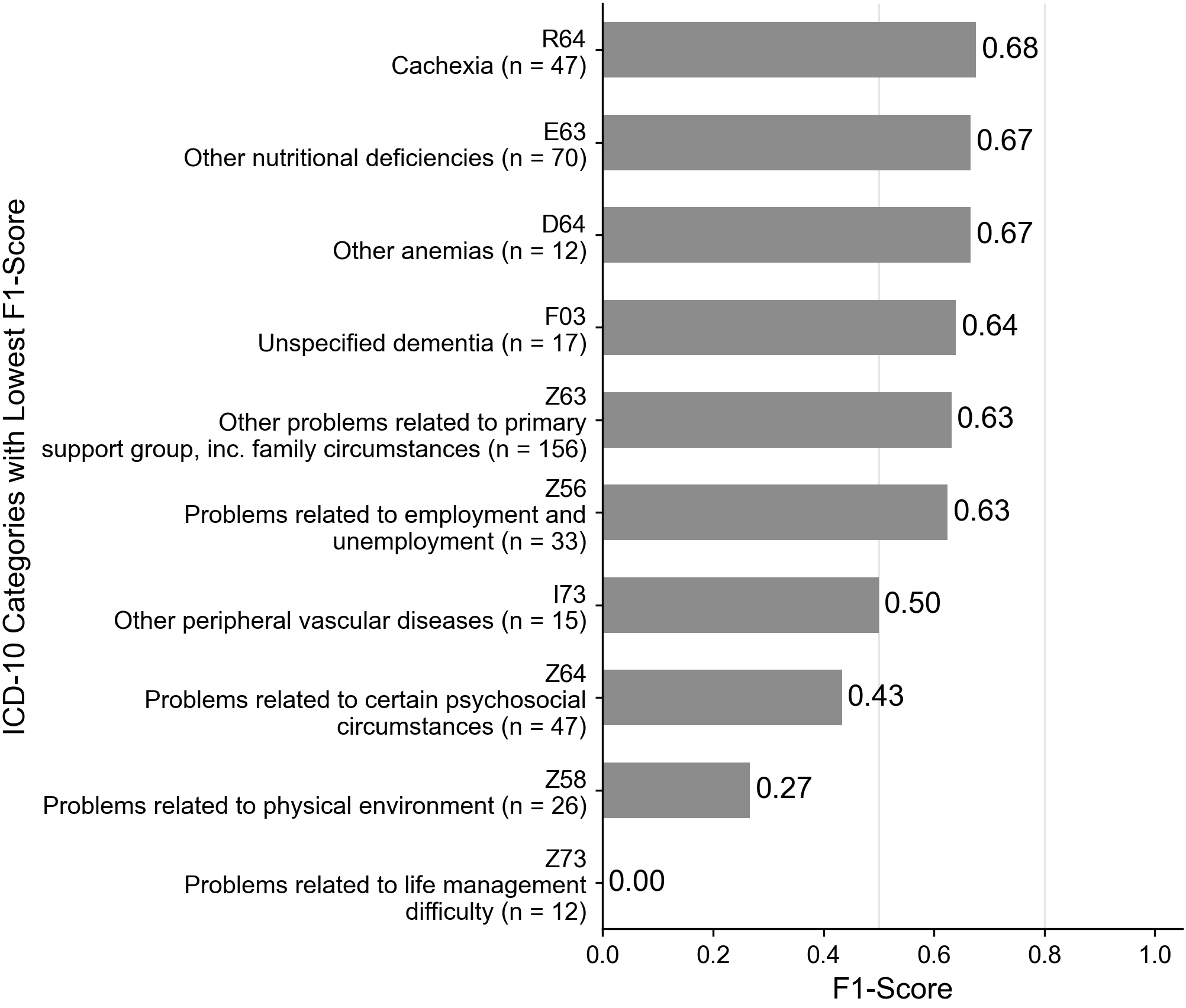}
\caption{}
\end{subfigure}
\begin{subfigure}{0.32\linewidth}
\centering
\includegraphics[width=\linewidth]{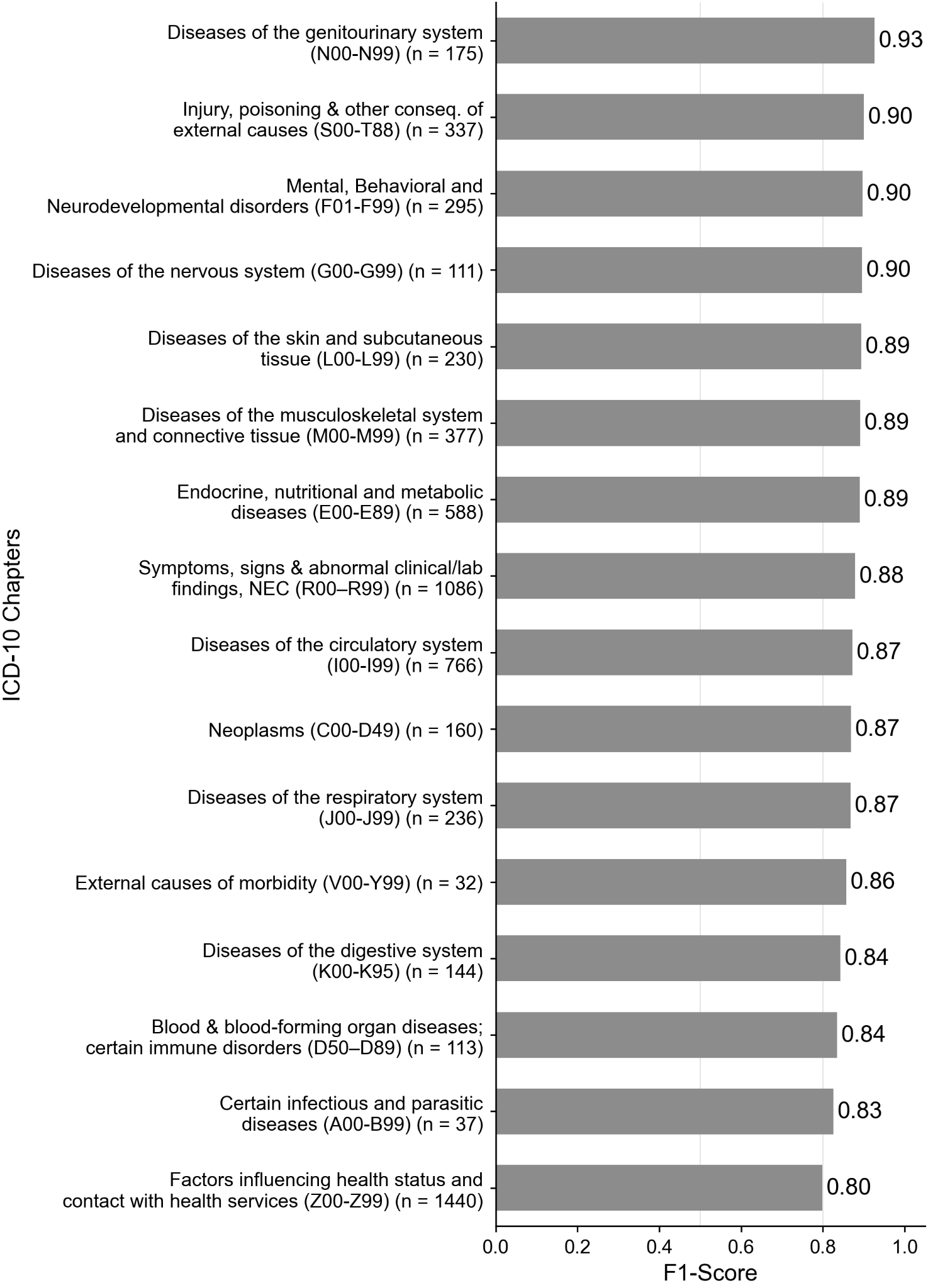}
\caption{}
\end{subfigure}
\par
\caption{Breakdown of ICD-10-CM coding performance across clinical domains and diagnostic groupings. (a) Weighted mean F1 score by clinical domain for Advanced Illness, Frailty, and Social Determinants of Health (SDoH). Error bars represent the weighted standard deviations. (b) Lowest-performing ICD-10-CM category-level codes with at least 10 evaluation cases. (c) Mean F1 score by ICD-10-CM chapter. Results are computed over 761 held-out synthetic clinical charts.}
\end{figure}

\begin{figure}[ht]
\centering
\includegraphics[width=0.9\linewidth]{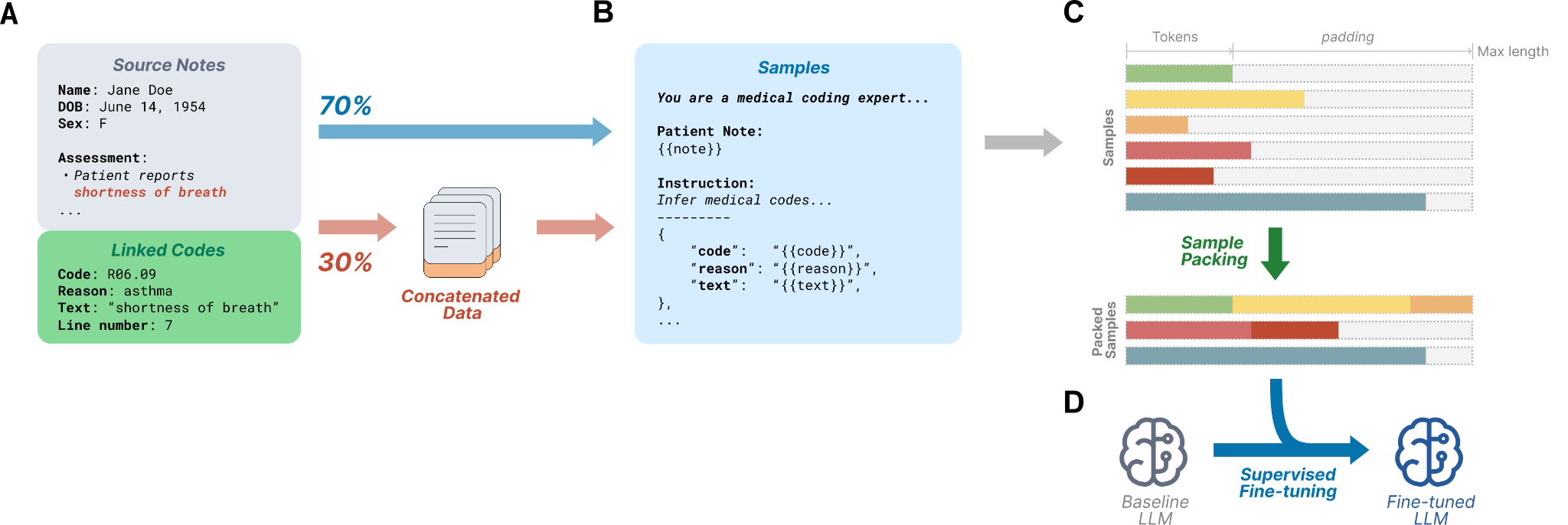}
\caption{Figure 4. Training workflow including data augmentation and sequence packing. A. Synthetic notes and linked ICD-10-CM and CPT labels are bundled. The training dataset is augmented by concatenating multiple notes to increase contextual difficulty and formatted using structured prompts. This amounts to 30\% of the original volume of notes. Data augmentation and packing are applied during training only. B. Samples are prepared in a format including system prompt, instruction, and output format. C. Samples are packed into fixed-length sequences to reduce padding and improve computational efficiency. D. The resulting sequences are used for supervised fine-tuning of the Llama-3-70B model.}
\end{figure}

\begin{figure}[ht]
\centering
\includegraphics[width=0.9\linewidth]{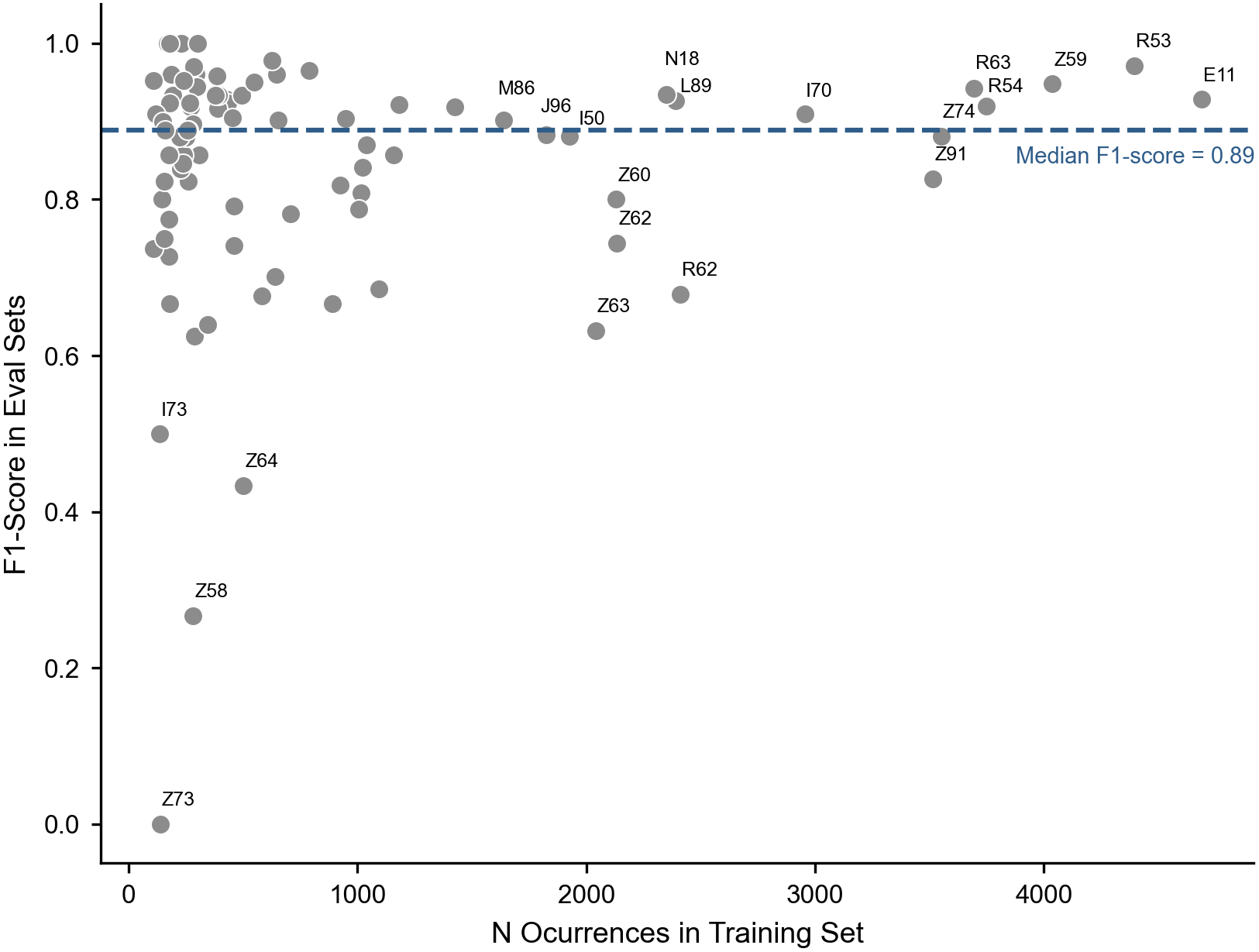}
\caption{Figure 5. Relationship between ICD-10-CM training frequency and evaluation performance. Scatter plot of category-level F1 score in the evaluation set versus the number of occurrences in the training data (categories with $\geq$10 evaluation cases). Low-frequency categories exhibit high variance in performance, while higher-frequency categories consistently achieve strong F1 scores, indicating a frequency threshold effect rather than a linear relationship.}
\end{figure}

\begin{figure}[ht]
\centering
\includegraphics[width=0.9\linewidth]{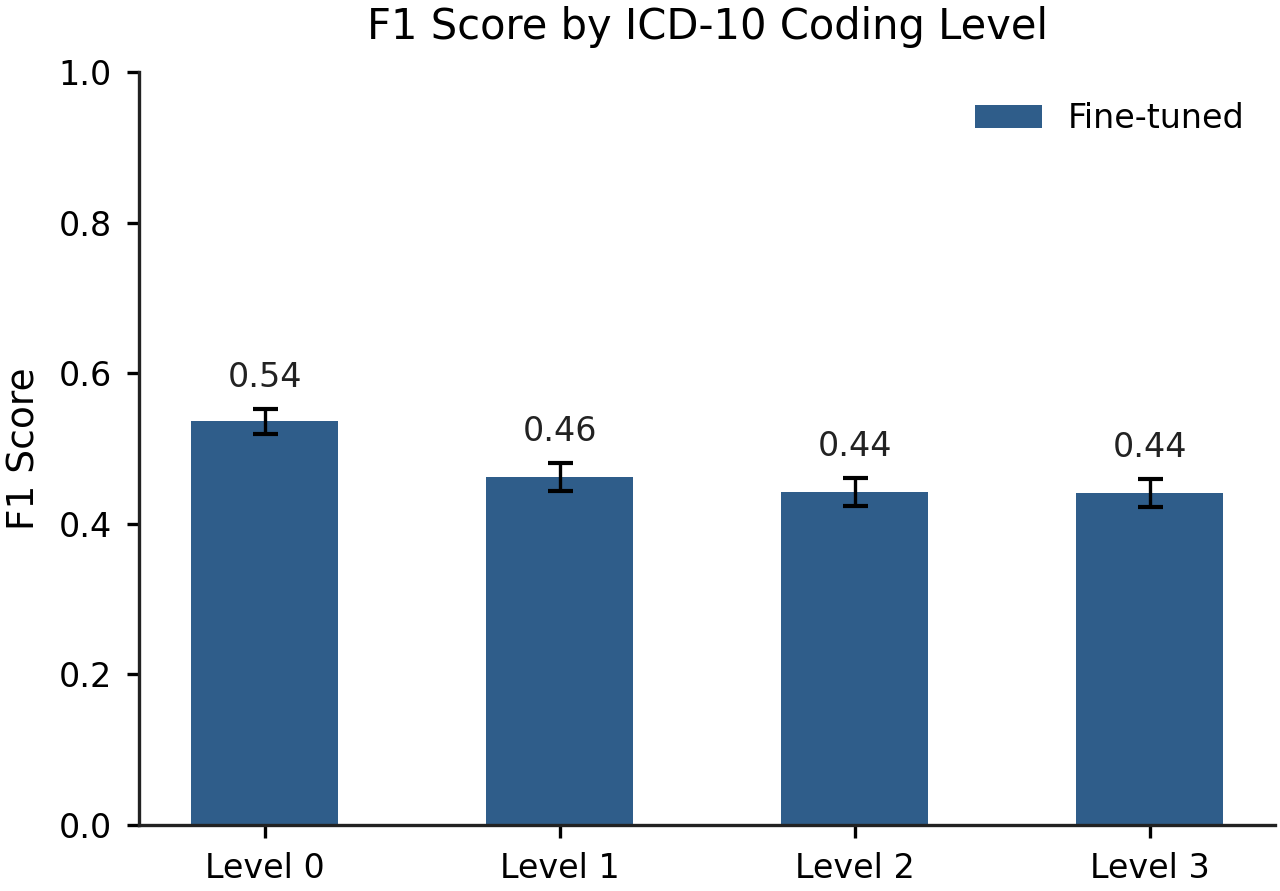}
\caption{Figure 6. Human expert evaluation of ICD-10-CM coding. Expert-validated ICD-10-CM coding accuracy on 100 synthetic charts.}
\end{figure}

\begin{table}[ht]
\centering
\caption{Table 1. ICD-10-CM coding performance by clinical domain. Weighted mean and standard deviation of precision, recall, and F1 score for Advanced Illness, Frailty, and Social Determinants of Health (SDoH). Metrics were first computed at the ICD-10 code level and then aggregated across codes using case-count--weights. Values reflect performance of the fine-tuned model on held-out synthetic evaluation data.}
\begin{tabular}{lccc}
\toprule
Category & Precision & Recall & F1 \\
\midrule
Advanced Illness (n=896) & \makecell{0.9008\\(0.1008)} & \makecell{0.8380\\(0.1184)} & \makecell{0.8630\\(0.1039)} \\
Frailty (n=2128) & \makecell{0.8727\\(0.1086)} & \makecell{0.8749\\(0.1303)} & \makecell{0.8727\\(0.1173)} \\
SDoH (n=1176) & \makecell{0.7716\\(0.1391)} & \makecell{0.7640\\(0.1578)} & \makecell{0.7668\\(0.1491)} \\
Combined (n=4200) & \makecell{0.8504\\(0.1268)} & \makecell{0.8360\\(0.1441)} & \makecell{0.8410\\(0.1329)} \\
Combined w/o SDoH (n=3024) & \makecell{0.8810\\(0.1071)} & \makecell{0.8639\\(0.1280)} & \makecell{0.8698\\(0.1136)} \\
\bottomrule
\end{tabular}
\end{table}

\begin{table}[htbp]
\centering
\caption{Performance on medical comprehension benchmarks before and after fine-tuning. Accuracy on standard medical question-answering and reasoning benchmarks for the baseline and fine-tuned Llama 3-70B models. Results indicate modest changes following fine-tuning, with no evidence of catastrophic degradation in general medical knowledge.}
\begin{tabular}{lccc}
\toprule
Benchmark & Baseline & Fine-tuned & Diff \\
\midrule
Average & .8286 & .7994 & -0.0292 \\
MedMCQA (acc) & .7196 & .6842 & -0.0354 \\
MedQA (acc) & .7879 & .7133 & -0.0746 \\
MMLU Anatomy & .8370 & .7852 & -0.0518 \\
MMLU Clinical Knowledge & .8415 & .8377 & -0.0038 \\
MMLU College Biology & .9167 & .9097 & -0.0070 \\
MMLU College Medicine & .7572 & .7514 & -0.0058 \\
MMLU Medical Genetics & .9000 & .8900 & -0.0100 \\
MMLU Prof Medicine & .9154 & .8493 & -0.0661 \\
PubMedQA & .7820 & .7740 & -0.0080 \\
\bottomrule
\end{tabular}
\end{table}

\begin{table}[ht]
\centering
\caption{Table 3. : Expert-validated ICD-10-CM coding performance by hierarchy level. Mean and standard error of precision, recall, and F1 score for the fine-tuned model with post-processing, evaluated against expert-accepted labels across 100 synthetic notes spanning Advanced Illness, Frailty, and Social Determinants of Health. Predictions were filtered to retain only domain-relevant codes prior to evaluation. ICD-10 hierarchy level ranges from Level 0 to Level 3. Metrics were aggregated at the chart level.}
\begin{tabular}{lccc}
\toprule
Category & Precision & Recall & F1 \\
\midrule
ICD-10 (Level 0) & \makecell{0.3963\\(0.0172)} & \makecell{0.9301\\(0.0160)} & \makecell{0.5365\\(0.0168)} \\
ICD-10 (Level 3) & \makecell{0.3153\\(0.0178)} & \makecell{0.8621\\(0.0208)} & \makecell{0.4416\\(0.0186)} \\
ICD-10 (Level 1) & \makecell{0.3309\\(0.0178)} & \makecell{0.8820\\(0.0199)} & \makecell{0.4618\\(0.0185)} \\
ICD-10 (Level 2) & \makecell{0.3157\\(0.0178)} & \makecell{0.8621\\(0.0208)} & \makecell{0.4422\\(0.0185)} \\
\bottomrule
\end{tabular}
\end{table}

\clearpage
\textbf{Author Contributions}:
\noindent G.K. and J.R. conceived and directed the study. J.C., A.C., S.G., and R.M. designed the synthetic data generation pipeline and coding policy framework. J.C. led synthetic data generation, with support from S.G., A.C., and R.M. A.C., A.S., and K.V. provided clinical coding expertise, developed gold-standard code sets, and validated outputs. A.S. and K.V. led human expert evaluation of model predictions, with support from A.C., R.M., and R.S. J.R., M.W., and Z.Y. led model fine-tuning, training infrastructure, and optimization on Snowflake. M.W. designed the fine-tuning pipeline, including data augmentation and packing strategies. P.W. led model evaluation, with contributions from I.C., R.M., J.C., and V.Z.V.V. on evaluation design and performance analysis. G.K. wrote the manuscript with input from all authors. All authors reviewed and approved the final manuscript.

\section{References}

\begingroup
\renewcommand{\section}[2]{}

\endgroup

\end{document}